\documentclass[lettersize,journal]{IEEEtran}
\usepackage{amsmath,amsfonts}
\usepackage{algorithmic}
\usepackage{algorithm}
\usepackage{array}
\usepackage[caption=false,font=normalsize,labelfont=sf,textfont=sf]{subfig}
\usepackage{textcomp}
\usepackage{stfloats}
\usepackage{url}
\usepackage{verbatim}
\usepackage{graphicx}
\usepackage{graphbox}
\usepackage{cite}
\usepackage{xcolor}

\DeclareMathOperator*{\supremum}{supremum}
\DeclareMathOperator*{\minimize}{minimize}

\begin{document}

\title{Embedded Object Detection and Mapping in Soft Materials Using Optical Tactile Sensing}

\author{Jose A. Solano-Castellanos$^{1}$, Won Kyung Do$^{1}$, and Monroe Kennedy III$^{1}$
\thanks{$^{1}$Authors are members of the \textbf{A}ssistive \textbf{R}obotics and \textbf{M}anipulation \textbf{Lab} (\textbf{ARMLab})  in the Mechanical Engineering Department at Stanford University, Stanford, CA 94305, USA. {\tt\small \{jsolanoc, wkdo, monroek\}@stanford.edu.}
This work is supported by the National Science Foundation under Grant 2142773. Project website with videos can be found here: https://sites.google.com/stanford.edu/embeddedobjectdetection/} }



\maketitle

\begin{abstract}
In this paper, we present a methodology that uses an optical tactile sensor for efficient tactile exploration of embedded objects within soft materials. The methodology consists of an exploration phase, where a probabilistic estimate of the location of the embedded objects is built using a Bayesian approach. The exploration phase is then followed by a mapping phase which exploits the probabilistic map to reconstruct the underlying topography of the workspace by sampling in more detail regions where there are expected to be embedded objects. To demonstrate the effectiveness of the method, we tested our approach on an experimental setup that consists of a series of quartz beads located underneath a polyethylene foam that prevents direct observation of the configuration and requires the use of tactile exploration to recover the location of the beads. We show the performance of our methodology using ten different configurations of the beads where the proposed approach is able to approximate the underlying configuration. We benchmark our results against a random sampling policy.
\end{abstract}



\section{Introduction}
\IEEEPARstart{T}{he} sense of touch is fundamental for us humans to safely and robustly interact with our environment, it is arguably the most informative sense that we possess, even more so than the sense of sight. With a single touch of our hand almost instantly we are able to know if an object is hot or cold, if its surface is smooth or rough, if it is hard or soft and we can even form a quick mental map about the orientation and shape of the object. Without having to stop reading this paper you can reach your pocket or the closest drawer with your hand and get an accurate inventory of what is inside.

Robotics is still missing the reliability that our skin provides us in our day-to-day tasks. In the last decade, a great effort has been dedicated towards providing robots the ability to sense the environment through touch, more specifically with the use of optical tactile sensors. Examples of such sensors are GelSight \cite{Li2014LocalizationSensing}, which is able to provide high-resolution relief maps of object surfaces; Soft-Bubble \cite{Alspach2019Soft-bubble:Manipulation} which has shown to be able to classify, estimate the pose and track objects; and DenseTact 2.0 \cite{Do2023DenseTactReconstruction}, capable of not only reconstruct the shape of the object but also provide a 6-axis wrench estimation.

Tactile information for robotic manipulation is particularly appealing when the sense of sight cannot provide reliable information to complete the task. Let us go back to the idea of forming an inventory of the objects that are in your pocket without having to take them out: the object of interest (rigid in nature) is occluded by the fabric (soft and deformable) and the quickest and often must intuitive way of figuring out what it is is by pinching and squeezing your fingers through the fabric. 

The implications of robots having this ability could be very significant. For industrial applications, robots would be able to classify packages at distribution centers without having to open them; for in-home care, robots that are equipped with this capacity could assist humans in retrieving objects hidden by blankets or within clothes in a more efficient and natural way. In medical applications, tactile sensation could aid physicians in assessing superficial anatomical structures providing reliable relief maps for further evaluation. Even for bionic prosthesis, forming a relief map and reproducing the haptic sensation on another part of the body of the user could partially retrieve this lost way of interaction with the world.

\begin{figure}[t!]
	\centering
	\includegraphics[width=3.45in]{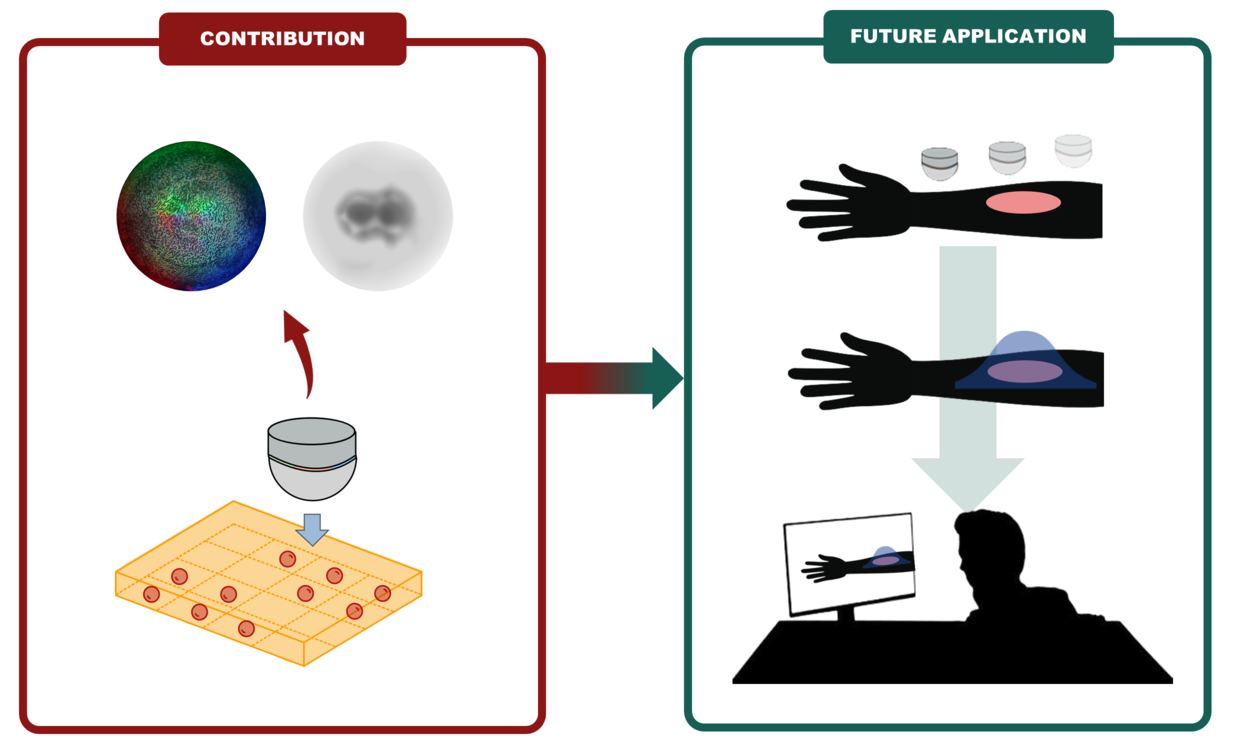}
  \caption{\textbf{Embedded Object Detection.} In this work we demonstrate the ability to detect and map objects embedded in a soft medium. This work has future applications to medical automated assistance for patient palpation.}
  	\label{fig:main}
      \vspace{-0.3cm}
\end{figure}

\subsection{Extracting Local Information Through Touch}

The use of optical tactile sensors to estimate mechanical properties and differentiate heterogeneous objects has been explored in the past. Yuan et al. \cite{Yuan2016EstimatingSensor}, \cite{Yuan2017Shape-independentSensor} used the GelSight sensor to estimate the hardness of everyday objects with different shapes, improving the object's recognition that could provide information for an appropriate grasping strategy. Yuan et al. \cite{Yuan2017ActiveLearning} also demonstrated the use of optical tactile sensors to actively recognize properties in clothing including characteristics such as thickness, fuzziness, softness, and durability and even estimate characteristics such as preferred washing methods or the season it is meant to be worn.

In the medical setting, Gwilliam et al. \cite{Gwilliam2010HumanTissue} proposed the use of a capacitive tactile sensor for the detection of lumps in soft tissues of variable size and depth. The sensor was compared with the human finger outperforming it in the detection of lumps, requiring lower indentation depths and pressures. Jia et al. \cite{Jia2013LumpSensor} later replicated the work using GelSight instead of the capacitive sensor, also showing better performance than the human finger.

\subsection{Active Exploration for Shape Reconstruction}

The above mentioned works show the efficacy of the use of optical tactile sensors to infer local properties of the objects. However, rich interactions with the environment require not only the capacity to detect local properties such as hardness, but also to gain global information of the object such as shape. One common choice for shape representation using haptic sensory information is through the use of implicit surfaces via Gaussian Processes (GP). Dragiev et al. \cite{Dragiev2011GaussianGrasping} show the performance of implicit surfaces GP shape estimation using tactile information from a seven-degrees-of-freedom hand.

Efficient shape reconstruction requires active and purposeful exploration of objects' surfaces. Martins \& Ferreira \cite{Martins2014TouchSurfaces} implemented a haptic exploration in simulation that allowed a humanoid robot to reconstruct the boundary between two different materials using a Bayesian model. Jamali et al. \cite{Jamali2016ActiveExploration} used a GP to probabilistically reconstruct the shape of the object driving the exploration based on the estimated boundaries of the object. A similar approach was shown to work by Yi et al. \cite{Yi2016ActiveProcesses}. A more recent work involving the use of GelSight was developed by Wang et al. \cite{Wang20183DPriors} integrating the use of visual and tactile exploration with learned shape priors for 3D shape reconstruction. Liu et al. \cite{Liu2023GelSightSensing} demonstrated the application of optical tactile sensing along the entire surface of a finger for a three-finger robotic hand which enabled the recognition of a set of rigid objects with a single grasp.

Merging both local and global information for shape reconstruction was recently explored by Zhao et al. \cite{Zhao2023FingerSLAM:Feedback} with FingerSLAM. This method uses both visual and tactile information to reconstruct the relief map by loop closure of the two sensing modalities. This method requires direct visualization of the surface and has only been tested on rigid objects.

As shown before, shape reconstruction has been extensively studied for rigid and homogeneous objects using tactile information. When the object is heterogeneous and/or deformable, previous work has been focused on gathering local information to obtain mechanical properties or estimate the presence of anomalies such as lumps in medical applications. To the best of our knowledge, simultaneous detection and mapping for heterogeneous and deformable objects has not been addressed in the past. This paper makes a significant contribution toward addressing this goal.

Our main contributions in this paper are: first, different from other papers where tactile sensors are used to estimate the local presence of hard embedded objects in soft materials, we generate a probabilistic map of the location of such objects using a Bayesian approach in a larger region; second, we use this probabilistic map to drive the exploration of the workspace and approximate the shape of the embedded objects (heterogeneous medium), different from other papers where such shape reconstruction is done on rigid objects (homogeneous medium); finally, our method is able to detect and approximate the shapes of closely embedded and discontinuous objects within the soft material, independent of their continuity, relying solely on their spatial grouping within the workspace. The paper is divided in the following sections: Section \ref{sec:Exploration and Mapping} describes the proposed framework which is divided into an exploration and a mapping phase, phases that are driven by a reduction of the uncertainty and the presence of embedded objects respectively; Section \ref{sec:Experimental Setup} describes the experimental setup that was used to show the efficacy of our method using ten different configurations of the embedded objects; Section \ref{sec:Results and Discussion} presents the empirical results obtained across the ten different configurations where we show that our framework outperforms a fully random policy on both the exploration and mapping phases.


\section{Exploration and Mapping}\label{sec:Exploration and Mapping}

\begin{figure*}[!t]
\centering
\includegraphics[width=0.825\textwidth]{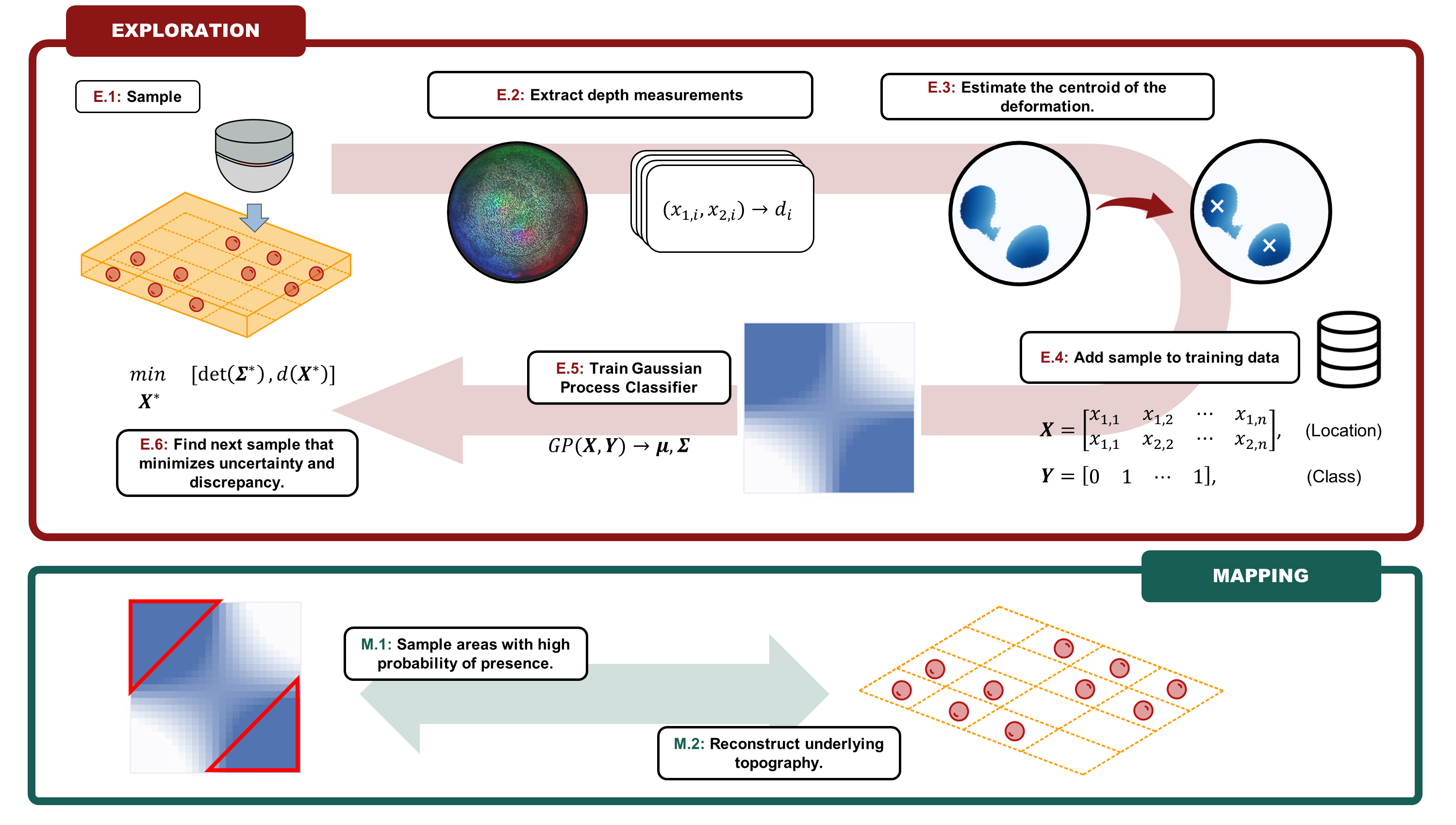}
\caption{Graphical representation of the proposed method. In the exploration phase (E.1 through E.6) the probable areas of hard embedded objects below the soft surface are estimated. In the mapping phase (M.1 and M.2) a more thorough interaction of such areas is conducted to approximate the underlying topology.}
\label{fig:ExplorationMapping}
\end{figure*}

We propose a framework for efficient tactile exploration and mapping for embedded rigid objects within matrices of soft materials using an optical tactile sensor. The method first generates a map that indicates the presence of hard objects below the surface (exploration), followed by a more thorough interaction in the areas of interest for an approximate reconstruction of the underlying topography (mapping). The method is tested in an experimental setup that involves a series of bead clusters in a planar array which is then covered by a polyethylene foam. Figure \ref{fig:ExplorationMapping} graphically represents the proposed method.

\begin{figure}[!t]
\centering
\includegraphics[width=\columnwidth]{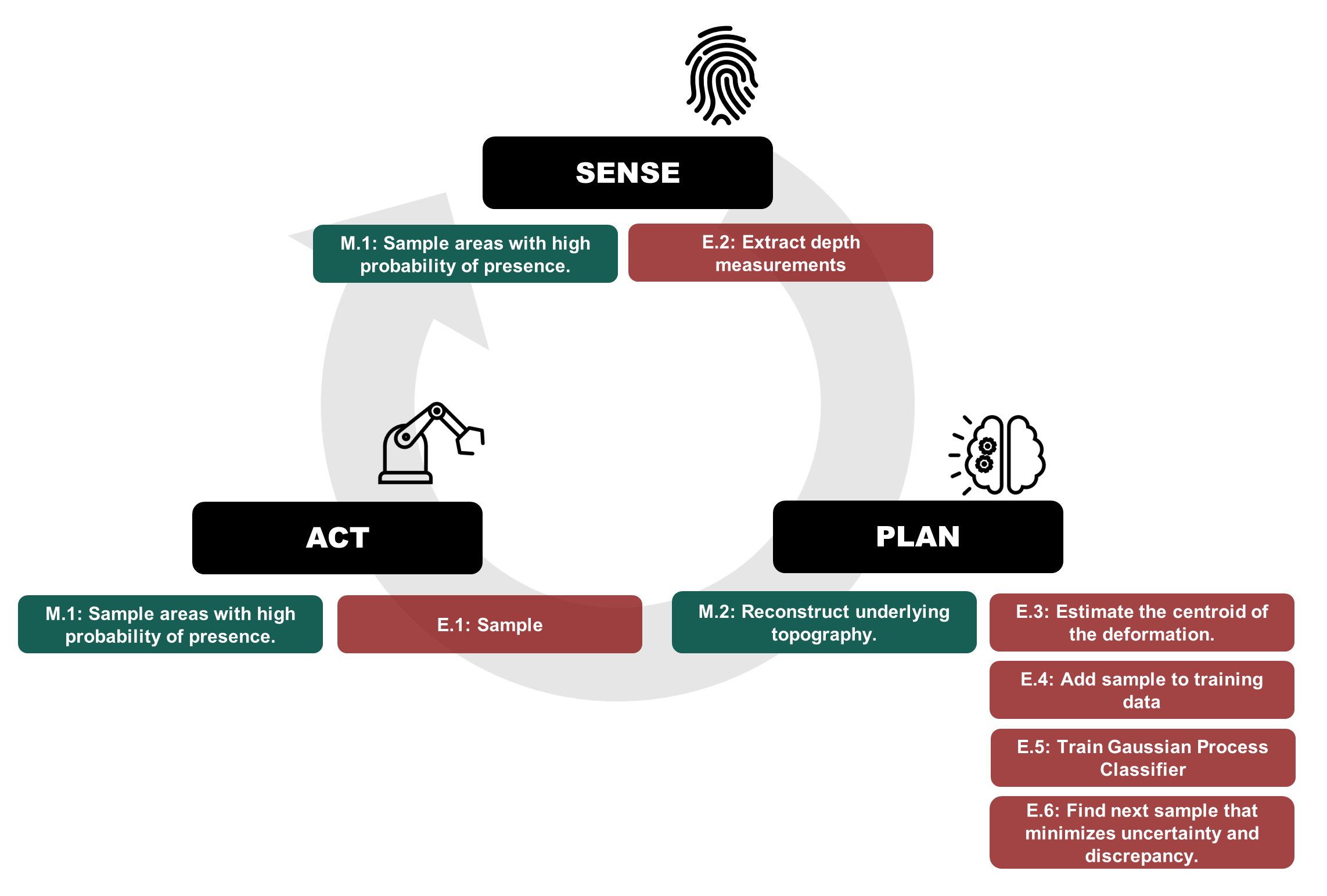}
\caption{Relation of the proposed method with the Sense-Plan-Act cycle of robotics.}
\label{fig:SeeThinkAct}
\end{figure}

It is important to highlight that our proposed method follows the Sense-Plan-Act control methodology of robotics as depicted in Figure \ref{fig:SeeThinkAct}.

\subsection{Exploration}

The exploration phase of the method, which provides a probabilistic map of the location of the hard objects below the soft surface starts by sampling the workspace (\textit{E.1} in Figure \ref{fig:ExplorationMapping}). Assuming that initially there is no prior information of the location of the embedded objects, the sampling can be performed using any desired strategy (e.g. randomized sampling). It is desired to get a comprehensive understanding of the workspace avoiding oversampling some regions of the workspace or leaving some regions unexplored. A sampling plan that encourages a diversity of samples and maximizes coverage of the workspace is called a \textit{space-filling} sampling plan. There exist many sampling plans that are considered space-filling, from which we have tested the Latin-hypercube sampling, Halton Sequence, and Sobol Sequences. Among these sampling strategies, including a randomized sampling plan, we found empirically that Sobol Sequences \cite{Sobol1967OnIntegrals} yields the best initialization for our experimental setup with eight initial samples to form the prior. Naturally, in case a prior is available, the use of a sampling strategy can be omitted.

Using DenseTact 2.0, each sample produces a $640 \times 640 \times 1$ depth
image (\textit{E.2} in Figure \ref{fig:ExplorationMapping}). The image is subsequently processed to check for the presence or absence of hard objects below the surface. To this end, the depth image obtained from the sample is compared against the image of the undeformed sensor, and a threshold $\varepsilon$ is applied: if the deformation (in absolute value) $|\delta|$ associated with a given pixel is larger than $\epsilon$, then the depth information is preserved, otherwise the deformation is set to zero. The value of $\varepsilon$ is closely related to the error of the sensor: if there is a deformation smaller than the error of the sensor, it is likely that it is noise from the sensor and not an actual deformation on the sensor. However, the presence of the soft material may affect the value of the threshold whose influence is difficult to quantify and deserves an analysis of its own. From our experiments, it was observed that the value of $\varepsilon$ needed to be set to 1.55 mm to produce a good performance of our method. The depth image is saved as a point cloud for later use in the mapping phase of the algorithm. 

Once the threshold was applied and deformation was detected in the depth image, the next step is to find the centroid of the indentations (\textit{E.3} in Figure \ref{fig:ExplorationMapping}). The depth image is run trough a Canny Edge Detector \cite{Canny1986ADetection} to find the contours of the clusters observed in the depth image. This allows us to form the convex hull of each cluster defined by the vertex set $\{\mathbf{v}_{i,1}, \cdots, \mathbf{v}_{i,n} \}$ and easily obtain the centroid $\bar {\mathbf{c}}_i = (x_i, \ y_i)$ of such set (\ref{eq:centroid}) which serves as an approximation of the centroid of the convex hull. In (\ref{eq:centroid}) $x_i$ and $y_i$ are the horizontal and vertical coordinates of the \textit{i-th} centroid respectively.

\begin{equation}
    \label{eq:centroid}
    \bar{\mathbf{c}}_i = \frac{1}{n}\sum_{j=i}^{n} \mathbf{v}_{i,j}
\end{equation}

This entire process is performed using the libraries {\tt{cv.Canny}}, {\tt{cv.findContours}}, {\tt{cv.convexHull}} and {\tt{cv.moments}} from OpenCV \cite{Bradski2000TheLibrary}.

If deformation is present on the depth image, the location of the centroid of each cluster is added to the training data with label $1$; if no deformation was present on the depth image, the location of the sample (center of the depth image) is added to the training data with label $0$ (\textit{E.4} in Figure \ref{fig:ExplorationMapping}). We have opted for the strategy involving the inclusion of a single training data point per cluster, as opposed to incorporating each individual pixel exhibiting deformation. This choice is driven by the need to address scenarios involving disjoint embedded objects and to facilitate predictions over larger areas of presence. By doing so, we prevent the Gaussian Process from optimizing for local information and encourage it to capture a more holistic understanding.

To generate the probabilistic map of the location of the embedded objects a Gaussian process for classification is used (\textit{E.5} in Figure \ref{fig:ExplorationMapping} and outlined in Algorithm \ref{alg:GP}). A Gaussian process $\mathcal{GP}(\mu(\cdot), \kappa(\cdot, \cdot))$ is a prior over functions defined by a mean function $\mu(\cdot)$ and a kernel function $\kappa(\cdot, \cdot)$. Different from a Gaussian process for regression, where we want to predict a continuous function value, in a Gaussian process for binary classification we care about a discrete variable $y^* \in \{1, 0\}$ at location $x^*$ given the observed classes $\mathbf{Y}$ at locations $\mathbf{X}$ using a predictive distribution $p(y^*=1 \ | \ x^*,\mathbf{X},\mathbf{Y})$ \cite{Rasmussen2006GaussianLearning.}, \cite{Bishop2006PatternStatistics}.

To compute such distribution, a latent function $\mathbf{f}(x)$ is introduced over which we perform a Gaussian process regression and then its output is passed through a logistic function (\ref{eq:logistic}), i.e. $p(y=1 \ | \ x ) = \sigma(\mathbf{f}(x))$. Please keep in mind that the latent function $\mathbf{f}$ serves as an intermediary variable within our method and does not possess a direct interpretable significance. 

\begin{equation}
    \sigma(z) = \left[1+\exp (-z) \right]^{-1}
    \label{eq:logistic}
\end{equation}

The probabilistic prediction is then defined as:

\begin{equation}
    p(y^*=1 \ | \ x^*,\mathbf{X},\mathbf{Y}) = \int \sigma(\mathbf{f}^*)p(f^* \ | \ x^*,\mathbf{X},\mathbf{Y}) d\mathbf{f}^*
\end{equation}

This integral is analytically intractable and is solved by first approximating the logit function with a probit function $\Phi(f^*)$ and the second term by using a Laplace approximation \cite{Rasmussen2006GaussianLearning.}. With these approximations, the integral is Gaussian and can be computed, therefore:

\begin{equation}
    p(\mathbf{f}^* \ | \ x^*,\mathbf{X},\mathbf{Y}) \approx \mathcal{N}(\mathbf{f}^* \ | \ \mathbf{\mu}, \mathbf{\Sigma})
\end{equation}
\begin{equation}
    \mathbf{\mu} = \mathbf{k}_*^T(\mathbf{Y}-\sigma(\hat{\mathbf{f}}))
    \label{eq:mean}
\end{equation}
\begin{equation}
    \mathbf{\Sigma} = \mathbf{k}_{**} -\mathbf{k}_*^T(\mathbf{W}^{-1}+\mathbf{K_f})^{-1}\mathbf{k}_*
    \label{eq:covariance}
\end{equation}

Where $\mathbf{k}_* = \kappa(\mathbf{X}, \mathbf{X}^*)$, $\mathbf{k}_{**} = \kappa(\mathbf{X}^*, \mathbf{X}^*)$, $\mathbf{W}$ is a diagonal matrix whose entries are $W_{ii} = \sigma(\text{f}_i)(1-\sigma(\text{f}_i))$, $\mathbf{K_f} = \kappa(\mathbf{X}, \mathbf{X}) + \nu\mathbf{I}$ and $\nu$ is a small value that provides numerical stability ($1\times10^{-5}$ in our case). The mean $\hat{\mathbf{f}}$ can be obtained iteratively, i.e. $\hat{\mathbf{f}} \leftarrow \mathbf{f}^{new}$ using (\ref{eq:f_new}).

\begin{equation}
    \label{eq:f_new}
    \mathbf{f}^{new} = \mathbf{K_f}(\mathbf{I}+\mathbf{W}\mathbf{K_f})^{-1}(\mathbf{Y}-\sigma(\mathbf{f}) + \mathbf{W}\mathbf{f})
\end{equation}

The predictive distribution can be approximated as follows \cite{Bishop2006PatternStatistics}:

\begin{equation}
    p(y^*=1 \ | \ x^*,\mathbf{X},\mathbf{Y}) \approx \sigma \left( \mathbf{\mu}\left( 1+\pi\mathbf{\Sigma}/8 \right)^{-1/2} \right)
    \label{eq:probability}
\end{equation}

We use a (isotropic) square exponential kernel (\ref{eq:kernel}) with parameters $\mathbf{\theta} = [\theta_1, \ \theta_2]$ which are optimized automatically by maximizing the log marginal likelihood $log \ p(\mathbf{Y} \ | \ \mathbf{\theta})$. In the experimental validation of our method (Section \ref{sec:Results and Discussion}) the values of $\mathbf{\theta}$ converged to $[30, \ 15]$.

\begin{equation}
    \kappa( \mathbf{x}_i, \mathbf{x}_j ) = \theta_2^2 \exp \left( -\frac{(\mathbf{x}_i - \mathbf{x}_j)^T(\mathbf{x}_i - \mathbf{x}_j)}{2\theta_1^2} \right)
    \label{eq:kernel}
\end{equation}

 \begin{algorithm}[H]
 \caption{Gaussian Process Classifier (\textit{E.5} in Figure \ref{fig:ExplorationMapping})}
 \label{alg:GP}
 \begin{algorithmic}[1]
 \renewcommand{\algorithmicrequire}{\textbf{Input:}}
 \renewcommand{\algorithmicensure}{\textbf{Output:}}
 \REQUIRE $\mathbf{X}$ (Seen Locations), $x^*$ (Unseen Location), $\mathbf{Y}$ (Class)
 \ENSURE $p(y^*=1 \ | \ \mathbf{X}^{*},\mathbf{Y})$ (Presence probability)
 \STATE Initialization: $\nu = 1 \times 10 ^{-5}$, $\mathbf{f} = \mathbf{0}$
 \STATE Compute: $\mathbf{K_f}$
 \FOR{j in max\_iter = 20}
    \STATE Compute: $\mathbf{W}$, $\mathbf{f}^{new} = \text{Eqn.} \ (\ref{eq:f_new})$ 
    \IF{$|\mathbf{f}^{new} - \mathbf{f}| < 1 \times 10^{-9}$}
    \STATE \textbf{break}
    \ENDIF
    \STATE $\mathbf{f} \leftarrow \mathbf{f}^{new}$
 \ENDFOR
 \STATE $\hat{\mathbf{f}} \leftarrow \mathbf{f}^{new}$
 \STATE Compute: $\mathbf{k}_*$, $\mathbf{k}_{**}$ and $\mathbf{W}$
 \STATE Compute: $\mathbf{\mu} = \text{Eqn.} \ (\ref{eq:mean})$, $\mathbf{\Sigma} = \text{Eqn.} \  (\ref{eq:covariance})$  
 \RETURN $p(y^*=1 \ | \ \mathbf{X}^{*},\mathbf{Y}) = \text{Eqn.} \ (\ref{eq:probability})$ 
 \end{algorithmic} 
 \end{algorithm}

With every sample, we aim to reduce the uncertainty in our probabilistic map. Since a Gaussian process also returns the confidence of the predicted label $y$ we can choose a sample $x^*$ that has the highest uncertainty as measured by the covariance matrix $\mathbf{\Sigma}$. However, in the tests conducted on real data, we found that this policy tends to oversample a region of the workspace before moving to others, which significantly delays the mapping phase of our algorithm. For this reason, we incorporate the discrepancy of the set of samples $\mathbf{X}^* = \{\mathbf{X} \ \cap \  x^* \}$ to decide the next sample. The discrepancy $d(\mathbf{X})$ is a criterion used to quantify the space-filling characteristics of a set of samples normalized over a unit hypercube and can be defined as follows \cite{Kochenderfer2019AlgorithmsOptimization}

\begin{equation}
    d(\mathbf{X}) = \supremum_\mathcal{H}\left| \frac{\#(\mathbf{X}\cap \mathcal{H})}{\#\mathbf{X}} - \lambda(\mathcal{\mathcal{H}}) \right|
\end{equation}

$\mathcal{H}$ is hyper-rectangular subset of the hypercube, $\#X$ is the number of samples, $\#(\mathbf{X}\cap \mathcal{H})$ is the number of samples that lie in $\mathcal{H}$, and $\lambda(\mathcal{\mathcal{H}})$ is the volume of $\mathcal{H}$. In general, the discrepancy is difficult to compute exactly and there exist approximate methods to do so. In our implementation we compute the discrepancy of the samples using the Centered Discrepancy (CD) method \cite{Zhou2013MixtureSets}. With this, we can find the next sample using both the uncertainty of the Gaussian process and the discrepancy of the samples (\textit{E.6} in Figure \ref{fig:ExplorationMapping}).

\begin{equation}
    \minimize_{\mathbf{X}^*} \quad [det \mathbf{\Sigma(\mathbf{X}^*)}, d(\mathbf{X}^*)]
\end{equation}

The selection of the next sample could be benefited by the presence of a human in the loop, which could be valuable in medical applications or in prostheses as outlined in the introduction. The entire loop (steps \textit{E.1} trough \textit{E.6}) is repeated to choose the next sample.

\begin{figure*}[!t]
    \centering
    \subfloat[]{\includegraphics[width=.335\textwidth]{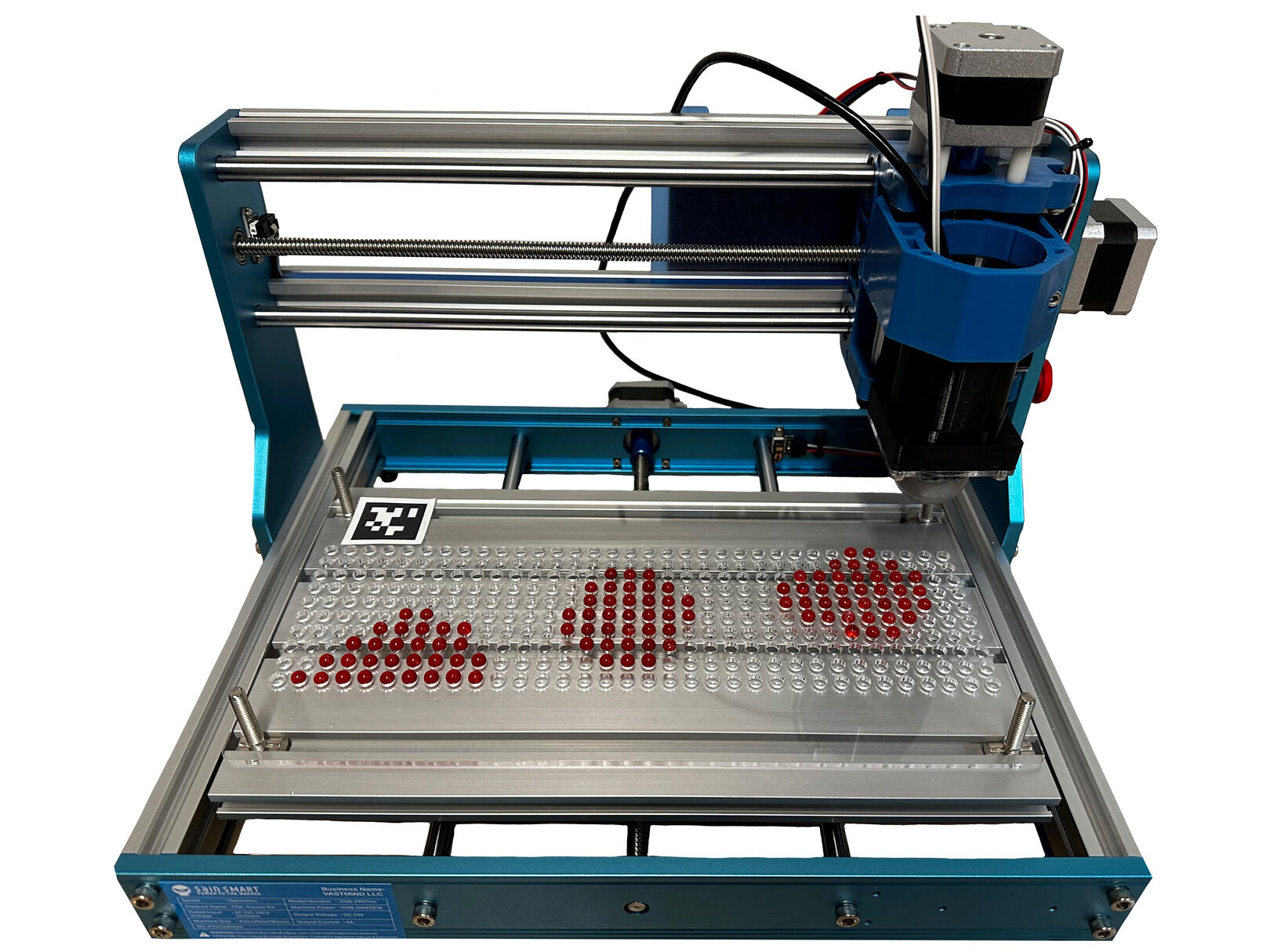}%
    \label{fig_first_case}}
    \hfil
    \subfloat[]{\includegraphics[width=.335\textwidth]{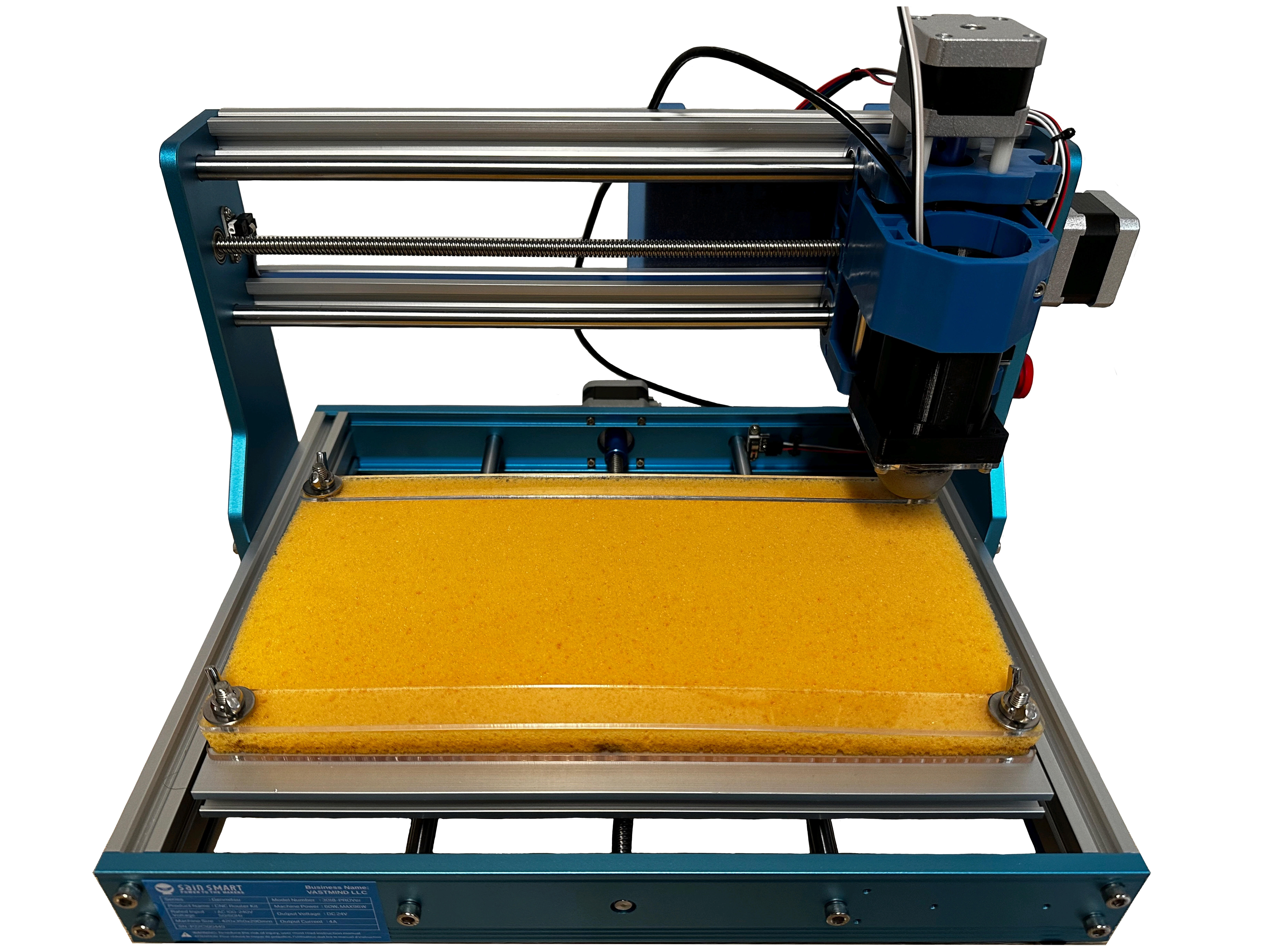}%
    \label{fig_first_case}}
    \hfil
    \subfloat[]{\includegraphics[width=.31\textwidth]{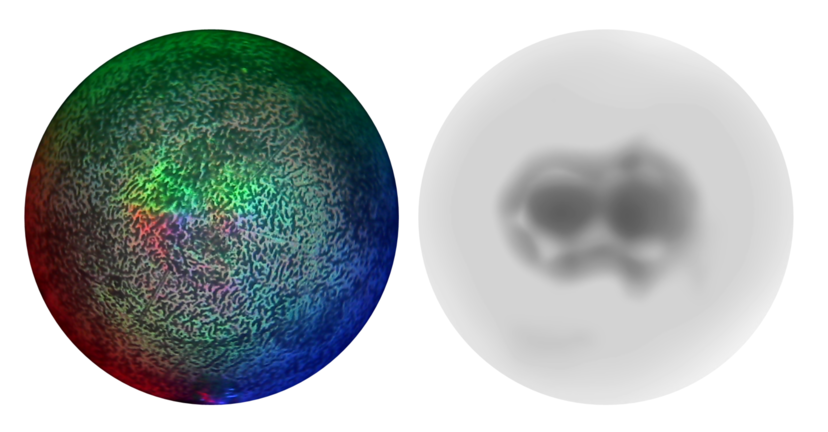}%
    \label{fig_first_case}}
    \caption{Experimental configuration for the detection and mapping of embedded rigid objects in soft material matrices. (a) A perforated acrylic sheet allows to accommodate quartz bead clusters in any desired configuration. (b) The beads are covered with a layer of polyethylene foam and secured in place with two longitudinal acrylic strips. (c) Side by side of the raw image from camera and the resulting depth image from DenseTact 2.0.}%
    \label{fig:ExprimentalSetup}%
\end{figure*}

\subsection{Mapping}

The mapping phase of the algorithm has a similar procedure to the exploration phase, however, the emphasis of this phase is to reconstruct the shape of the embedded objects rather than forming the probabilistic map of their location. Because of this, we sample the areas with a high probability of presence according to the Gaussian process (\textit{M.1} in Figure \ref{fig:ExplorationMapping}) following the same loop as in the exploration phase. The difference now is that we aim to sample areas with a high probability of presence rather than reduce the uncertainty in the workspace. The sampling strategy is then modified in step \textit{E.6} as:

\begin{equation}
    \minimize_{\mathbf{X}^*} \quad [-p(y^*=1 \ | \ \mathbf{X}^*,\mathbf{Y}), d(\mathbf{X}^*)]
\end{equation}

With each additional sample, we continue training the Gaussian process as well as reconstructing the underlying topography of the workspace (\textit{M.2} in Figure \ref{fig:ExplorationMapping}) which can then be visualized as a point cloud and further processing can be done if needed.


\section{Experimental Setup}\label{sec:Experimental Setup}

The experimental setup is presented in Fig. \ref{fig:ExprimentalSetup} where DenseTact 2.0 is attached to a desktop computer numerical control (CNC) machine for data collection. A perforated acrylic sheet with 5.5-millimeter holes in a 10 $\times$ 34 array with a center distance of 8.5 millimeters both vertically and horizontally, provides an easy to reconfigure test bed Fig. \ref{fig:ExprimentalSetup}(a). The perforated acrylic sheet allows us to locate quartz beads of 6 millimeters of diameter in the desired configurations for our experimental validation of the framework described in section \ref{sec:Exploration and Mapping}. This array provides an effective area of 90.4 $\times$ 300 millimeters.

Once the beads have been placed in the desired locations, the array is covered with a sheet of polyethylene foam with a thickness of half of an inch. The foam is secured in place with two longitudinal acrylic strips that are clamped to the CNC bed as shown in Fig. \ref{fig:ExprimentalSetup}(b). 

\begin{figure}[!t]
\centering
\includegraphics[width=\columnwidth]{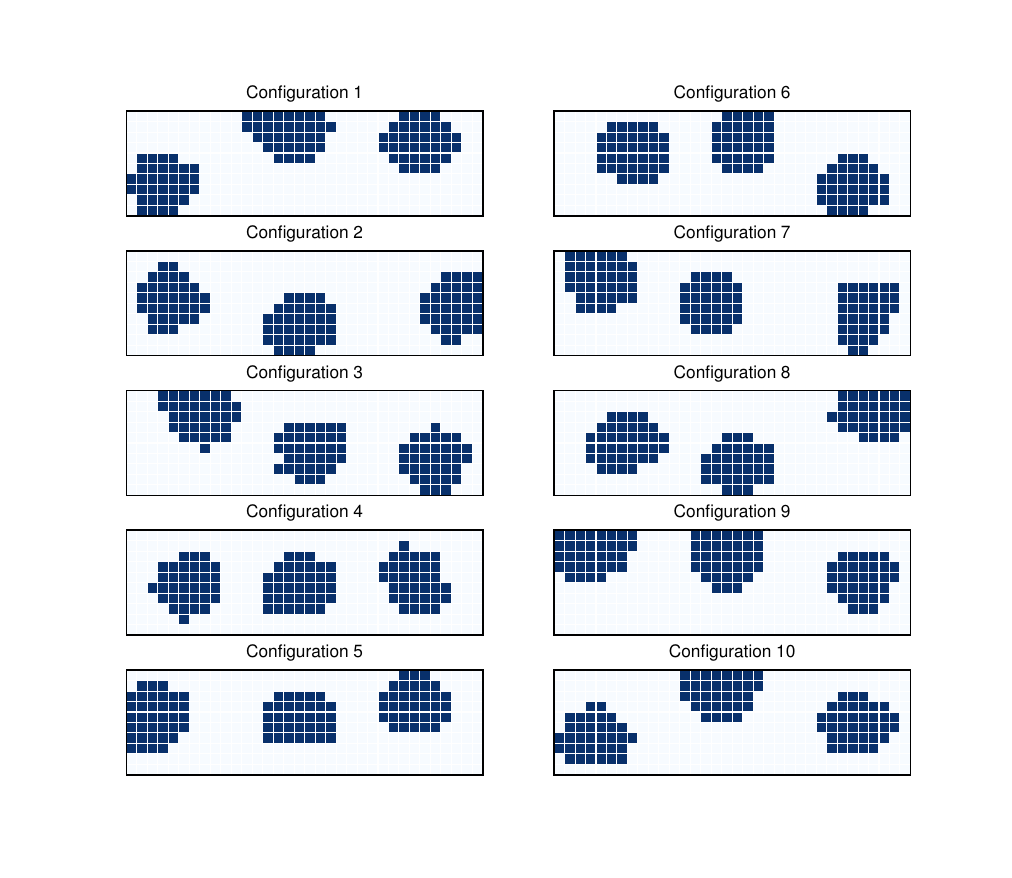}
\caption{Configurations of the quartz beads below the foam for the evaluation of the performance of the proposed method}
\label{fig:Configurations}
\end{figure}

For the evaluation of the performance of the framework, we evaluated ten different configurations on the beads on the acrylic sheet (Figure \ref{fig:Configurations}). Due to the dimensions of the workspace, each configuration was comprised of three bead clusters, each cluster with thirty-four beads on average. All the data was collected at the same depth of 12 millimeters below the surface of the foam. The beads are dyed red to be identified with a computer vision program, that together with an AprilTag \cite{Olson2011AprilTag:System} allows us to easily reconstruct the configuration for evaluation of the performance of our framework.

We are interested in reconstructing the underlying shape of the embedded objects. To accomplish this, it is necessary for the soft material to have relatively low stiffness against the sensor. In our case, DenseTact 2.0 has a Shore 00 hardness of 49.7, while the polyethylene foam has a Shore 00 hardness of 24.7 when pressed seven millimeters. For this reason, we limited our experiments to a single type of soft material. However, it is important to highlight that with a stiffer material, it is still possible to apply the exploration phase of our method as long as the sensor is able to detect the presence of the embedded materials as in \cite{Gwilliam2010HumanTissue} and \cite{Jia2013LumpSensor}.


\section{Results and Discussion}\label{sec:Results and Discussion}

\begin{figure*}[!t]
    \centering
    \subfloat[]{\includegraphics[width=.48\textwidth]{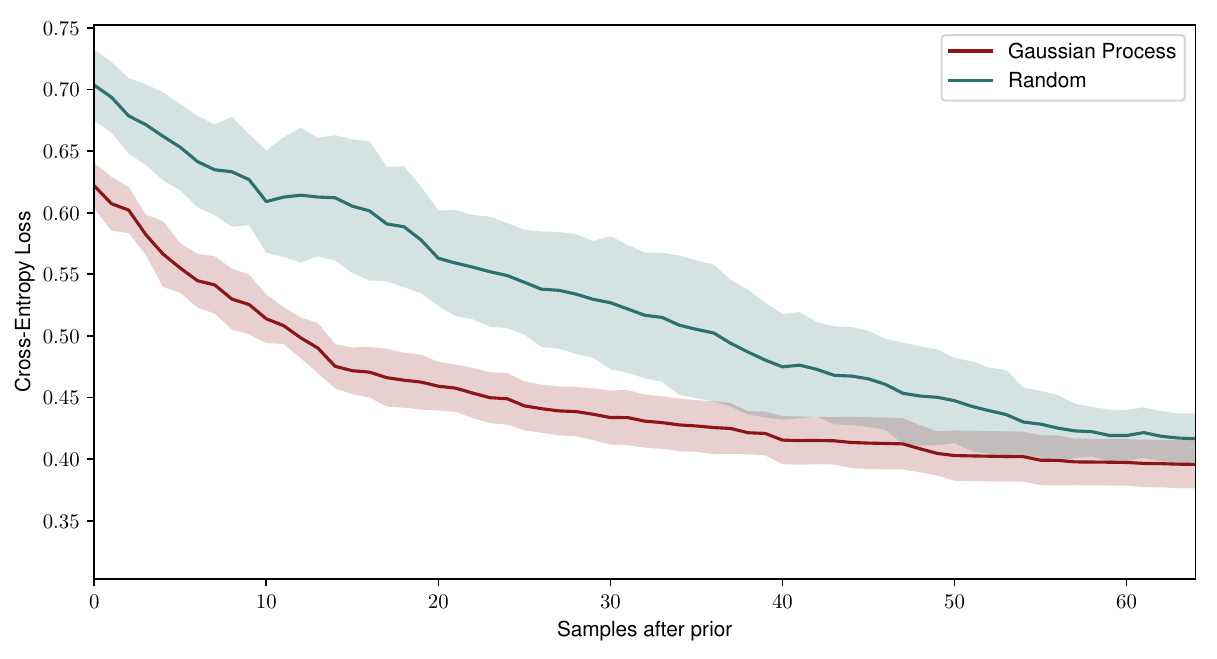}%
    \label{fig_first_case}}
    \hfil
    \subfloat[]{\includegraphics[width=.48\textwidth]{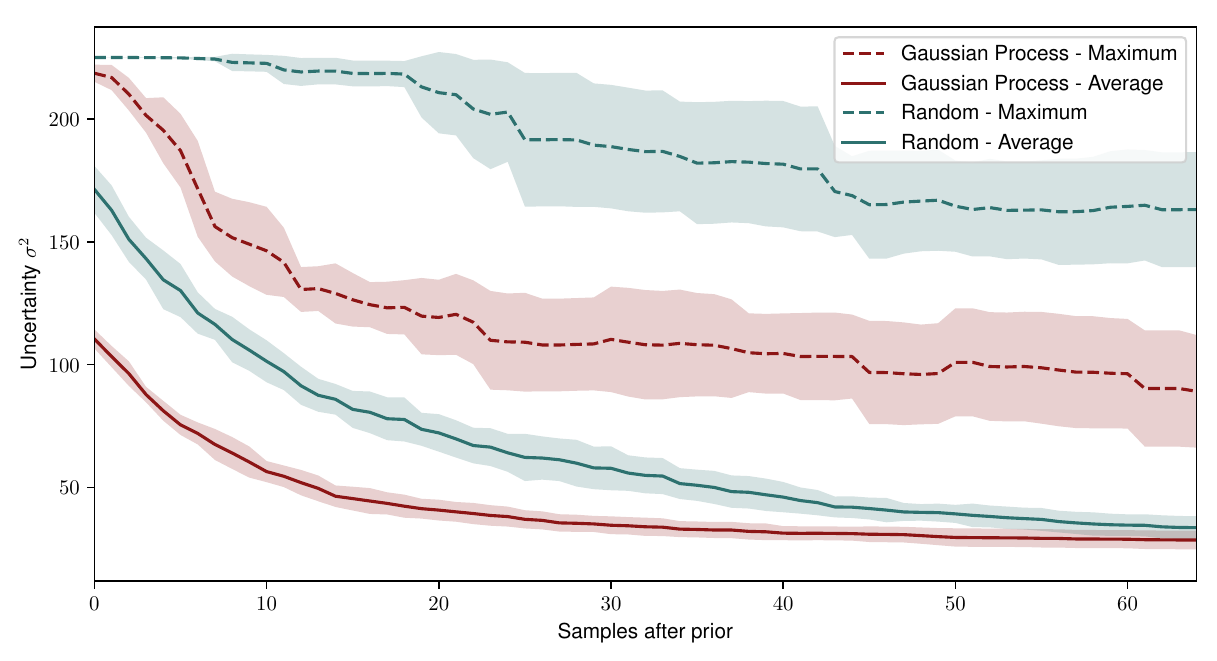}%
    \label{fig_first_case}}
    \caption{Performance of the exploration phase showing the average over the ten configurations shown in Figure \ref{fig:Configurations} with a solid\slash dashed lines and the shaded region indicating the corresponding standard deviation. (a) Shows the Cross-Entropy loss of our proposed method in red and a fully random policy in green. (b) Shows the uncertainty, both maximum (dashed line) and average (solid line), for our proposed method and a fully random policy.}%
    \label{fig:ExplorationPerformance}%
\end{figure*}

\begin{figure*}[!t]
\centering
\subfloat[]{\includegraphics[width=0.98\textwidth]{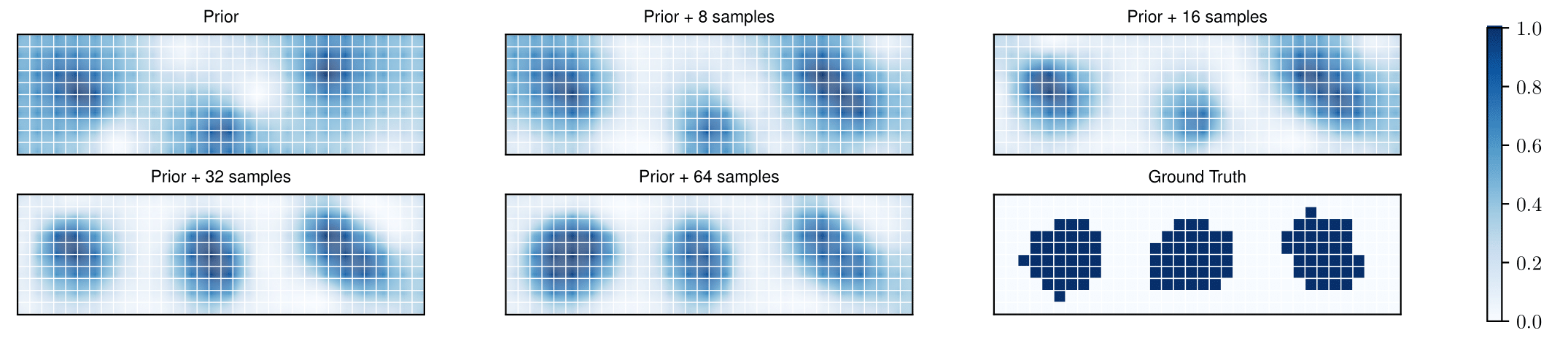}%
\label{fig_first_case}}
\hfil
\subfloat[]{\includegraphics[width=0.98\textwidth]{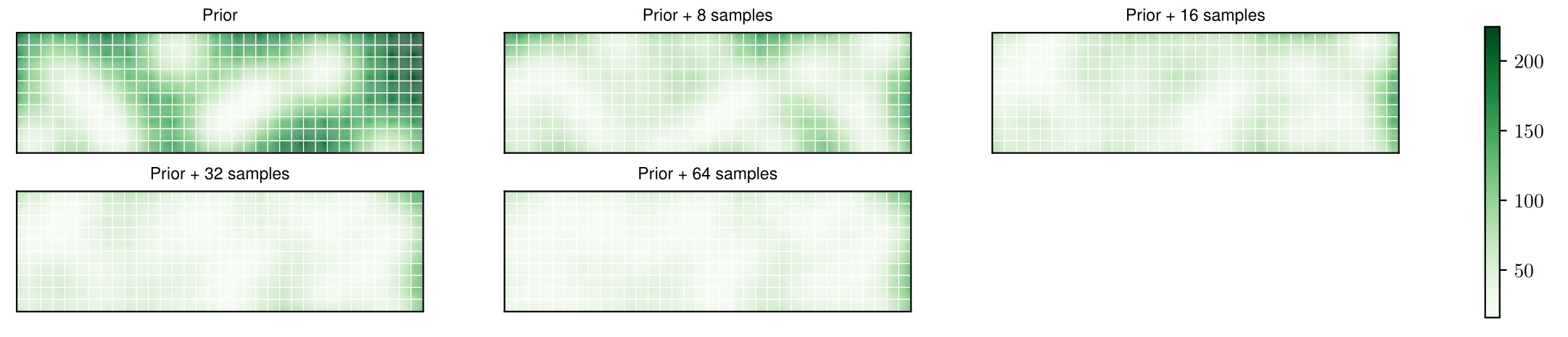}%
\label{fig_first_case}}
\caption{Example of the progression of the exploration phase of the proposed method (\textit{E.4} in Figure \ref{fig:ExplorationMapping}). (a) Shows the probabilistic map, given by equation (\ref{eq:probability}), obtained after the prior and additional eight, sixteen, thirty-two, and sixty-four additional samples as well as the ground truth. Darker regions denote a high probability of presence, while lighter regions denote a low probability of presence. (b) Shows the associated variance of the plots shown in (a) given by equation (\ref{eq:covariance}). Darker regions denote high uncertainty, while lighter regions denote low uncertainty.}%
\label{fig:Example_Exploration}
\end{figure*}

Figure \ref{fig:ExplorationPerformance} shows the performance of the exploration phase with up to sixty-four samples after the prior. The results are benchmarked against a fully random policy where the prior is also formed with eight randomly chosen samples. Regarding this random policy, we implemented a discretization of the workspace at 2.5 mm intervals in both vertical and horizontal directions. This discretized grid serves as the basis for randomly selecting samples from across the workspace. Because the main result of the exploration phase is the probabilistic map of the presence or absence of embedded objects (categorical variable), we evaluate the performance using the Cross-Entropy loss (\ref{eq:CE_Loss}) with a density of 0.5 millimeters both horizontally and vertically:

\begin{multline}
    \label{eq:CE_Loss}
    \mathcal{L}_{CE} = -\frac{1}{N}\sum_{i=1}^N \left[ \ y_i \ log \ p(y_i=1 \ | \ \mathbf{X},\mathbf{Y}) \ + \right. \\ 
    \left. (1 - y_i) \ log \ p(y_i= 0 \ | \ \mathbf{X},\mathbf{Y}) \ \right]
\end{multline}

As can be seen in Figure \ref{fig:ExplorationPerformance}(a) the proposed method has a better performance as compared with the fully random policy. Our proposed method has two distinct behaviors: before 14 to 16 samples after the prior, the loss decreases rapidly; after 14 to 16 samples after the prior, the loss continues decreasing but with a slower rate. As shown in Figure \ref{fig:Example_Exploration}, with sixteen samples after the prior, the probabilistic map has capture the main characteristic of the true distribution of the embedded objects. Different from our proposed method, the random policy more or less decreases at the same rate up to sixty additional samples after the prior. Eventually with enough samples, both strategies reach a similar loss. However, it is important to highlight that our proposed method has a better consistency as reflected in a tighter standard deviation.

In terms of uncertainty, Figure \ref{fig:ExplorationPerformance}(b), there is a similar behavior to the loss: before 14 to 16 samples there is a rapid decrease of uncertainty, both the maximum and average value, followed by a slower rate of decrease. As shown in the figure, both maximum and average uncertainty are lower in our method compared to the random strategy.

Since we are interested in forming a quick idea of the location of the embedded objects to sample the areas with a high probability for the mapping phase, it is valuable that our method can capture the main features with approximately 14 to 16 samples, with a lower uncertainty that a fully random sampling strategy. Given the observed trend in Figure \ref{fig:ExplorationPerformance}, we choose to have an exploration phase of 16 samples (or equivalently a Cross-Entropy loss of around 0.45 for the given size of the workspace) after the prior before proceeding to the mapping phase of our method.

For the mapping phase, the evaluation of the performance is accomplished by using the Mean Squared Error (MSE) loss (\ref{eq:L2_Loss}) between the ground truth $z$ and the height reported by the sensor $\Tilde{z}$, again with a density of 0.5 millimeters both horizontally and vertically:

\begin{equation}
    \label{eq:L2_Loss}
    \mathcal{L}_{MSE} = \frac{1}{N} \sum_{i=1}^N (z_i - \tilde{z}_i)^2
\end{equation}

\begin{figure*}[!t]
\centering
\includegraphics[width=0.98\textwidth]{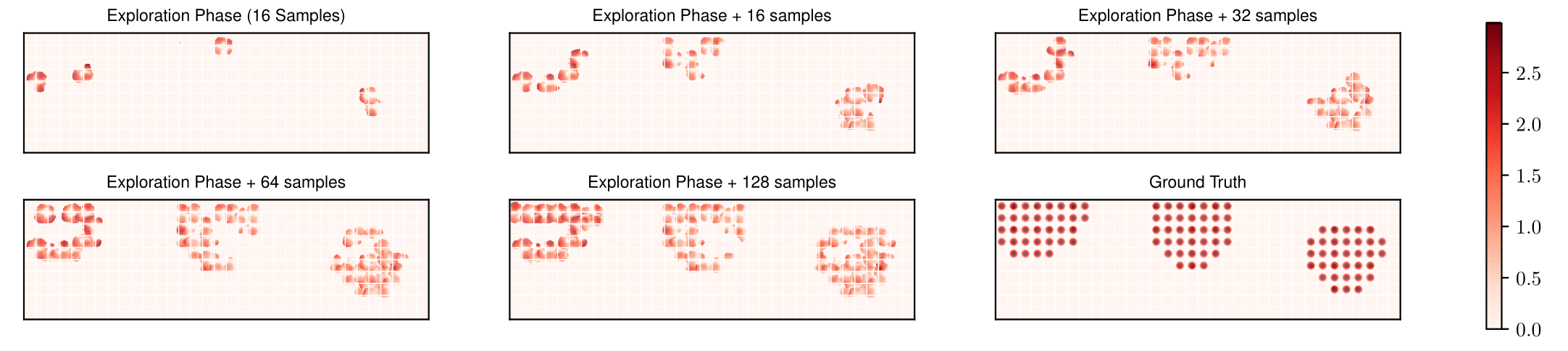}
\caption{Example of the progression of the mapping phase of the proposed method (\textit{M.2} in Figure \ref{fig:ExplorationMapping}). The image shows the approximate topology obtained after the exploration phase and additional sixteen, thirty-two, sixty-four, and one hundred and twenty-eight additional samples as well as the ground truth. The scale on the right is given in millimeters.}
\label{fig:Example_Mapping}
\end{figure*}

\begin{figure}[!t]
\centering
\includegraphics[width=0.95\columnwidth]{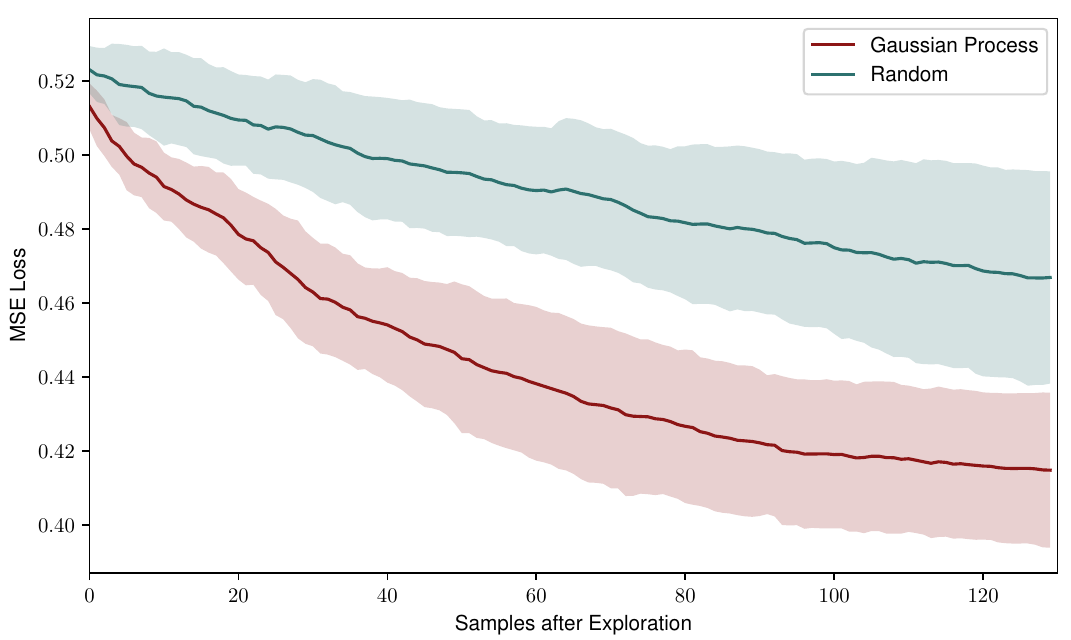}
\caption{Performance of the mapping phase showing the average over the ten configurations shown in Figure \ref{fig:Configurations} with a solid line and the shaded region indicating the corresponding standard deviation.}
\label{fig:MappingPerformance}
\end{figure}

As seen in Figure \ref{fig:MappingPerformance} the proposed method has a better performance as compared with the fully random policy: throughout the one hundred and twenty-eight samples the MSE error is consistently lower for our proposed method. This is expected since using a random policy it is possible that regions that do not have any embedded objects are sampled, especially in this case where the majority of the workspace does not have embedded objects. Our method in contrast does focus its sampling efforts on the areas that the exploration phase estimates are embedded objects. Finally, although not as noticeable as in the exploration phase, our method is more consistent than the fully random policy as depicted by the tighter standard deviation in Figure \ref{fig:MappingPerformance}. Figure \ref{fig:Example_Mapping} provides and example of the progression of the shape reconstruction of the embedded objects below the polyethylene foam.

The performance of our method could be improved by incorporating local information about stress and force. Although DenseTact 2.0 provides a 6-axis wrench estimation over the entire sensor, when we incorporated this information in our method, due to some inconsistency and high variance of the measurements (either from issues in the calibration or undesired interactions between the soft material and the sensor) our method had a tendency to substantially favor depth information and for this reason, we dispensed with strength information.


\section{Conclusion}\label{sec:Conclusion}

In this paper, we have proposed a method for the exploration and mapping for heterogeneous materials, where hard embedded objects are present within a matrix of a soft and deformable material, situation in which sight cannot provide reliable or any kind of information and tactile sensing becomes the main mechanism of interaction. Our method has been motivated in applications such as package sorting, medical diagnosis, and restoration of haptic sensation in individuals with prostheses.

The proposed framework is divided into two main phases. In the exploration phase, where a probabilistic map is generated by sampling the workspace and using a Gaussian process for classification we estimate the probability of the presence or absence of the hard embedded objects. The mapping phase exploits the probabilistic map generated by the exploration phase by sampling the areas of high probability of presence to reconstruct the underlying shape of the hard objects. We also present a strategy to obtain a prior of the location of the embedded objects using Sobol sequences when such prior is not available.

We validate our approach using an experimental setup that located a series of quartz beads in clusters underneath a polyethylene foam, and by using the optical tactile sensor DenseTact 2.0 together with the aid of a computer numerical control (CNC) machine we sample the workspace. 

Our empirical results show that our method outperforms a fully random policy on both the exploration and mapping phase. In the exploration phase, our method presents two distinct behaviors, before 14 to 16 samples after the prior there is a rapid decrease in both uncertainty and Cross-Entropy loss, followed by a continuous decrease of both the uncertainty and loss with a slower rate. It is valuable to see the rapid decrease in both loss and uncertainty with few samples to be able to transition to the mapping phase and focus the attention of the samples on the shape reconstruction aspect of the method. For the given size of the workspace, 16 samples (or equivalently a Cross-Entropy loss of around 0.45) were chosen to be the number of samples before switching to the mapping phase where we also demonstrate that our method outperforms the fully random policy. On both exploration and mapping phases, our proposed method presents a better consistency as compared to the random policy, shown by the smaller standard deviation across the ten different bead configurations.

Immediate extensions include the post-processing of the reconstructed shape depending on the application: for item recognition in packages and home assistance, the reconstructed shape can be passed through a convolutional neural network to identify the different objects; for medical applications, the reconstructed shape can be displayed to the physician using virtual reality for assessment; and or haptic sensation, the reconstructed shape can be reproduced on another part of the prosthesis owner's body. Future work could include the sampling and reconstruction of three-dimensional and\slash or curved surfaces and incorporate stress and force measurements into the method to obtain additional information of the embedded objects.




\end{document}